\documentclass{article}
\usepackage[utf8]{inputenc}
\usepackage[ruled,vlined]{algorithm2e}
\usepackage{graphicx}
\usepackage{authblk}
\usepackage{booktabs}
\usepackage[numbers]{natbib}
\usepackage{url}
\usepackage{subcaption}
\usepackage[table,xcdraw]{xcolor}
\usepackage [autostyle, english = american]{csquotes}
\MakeOuterQuote{"}
\usepackage[capitalise]{cleveref}

\title{An Adaptive and Near Parameter-free Evolutionary Computation Approach Towards True Automation in AutoML}

\begin{document}

\author{Benjamin Patrick Evans}
\author{Bing Xue}
\author{Mengjie Zhang}
\affil{School of Engineering and Computer Science \\ Victoria University of Wellington \\ New Zealand}
\affil{\textit {\{benjamin.evans,bing.xue,mengjie.zhang\}@ecs.vuw.ac.nz}}
\date{}

\maketitle

\begin{abstract}
    A common claim of evolutionary computation methods is that they can achieve good results without the need for human intervention. However, one criticism of this is that there are still hyperparameters which must be tuned in order to achieve good performance. In this work, we propose a near "parameter-free" genetic programming approach, which adapts the hyperparameter values throughout evolution without ever needing to be specified manually. We apply this to the area of automated machine learning (by extending TPOT), to produce pipelines which can effectively be claimed to be free from human input, and show that the results are competitive with existing state-of-the-art which use hand-selected hyperparameter values. Pipelines begin with a randomly chosen estimator and evolve to competitive pipelines automatically. This work moves towards a truly automatic approach to AutoML.
\end{abstract}

\section{Introduction}

In recent years, machine learning has made its way into many application areas, which has attracted a wide variety of interest from many users from outside the machine learning world. This demand for machine learning has spurred the area of automated machine learning (AutoML), which aims to make machine learning accessible to non-experts \cite{gijsbers2019gama}, or allows experts to focus on other aspects of the machine learning process rather than pipeline design \cite{OlsonGECCO2016}. 

However, while two of the goals of AutoML are automation and ease of use, most current state-of-the-art methods become a new optimisation problem themselves: rather than searching for pipelines, one must search for appropriate hyperparameters. Granted, this is a simpler search space than the original one, but is still an undesirable property and prevents true human-free automation.

In this work, we aim to overcome the aforementioned limitation by proposing an adaptive evolutionary computation method, which requires no specification of evolutionary hyperparameters and finds good performing pipelines automatically. The proposed method starts with a single randomly estimator with no preprocessing steps and automatically evolves to a well-performing pipeline without requiring any hyperparameters to be set. The entire evolutionary process is achieved without any human configuration.

The major contribution is a truly automated approach to AutoML, based on a near parameter-free evolutionary computation approach, which is able to rival human selected hyperparameters for current state-of-the-art systems.

\section{Background and Related Work}\label{secBackground}

There are three related areas here: parameter-free optimisation, adaptive hyperparameters, and automated machine learning. Unfortunately, there are no current works on parameter-free (or adaptive) automated machine learning, so we look at the areas separately in this section, before introducing the first adaptive parameter-free AutoML approach in \cref{secProposed}.

\subsection{Automated Machine Learning}
The goal of AutoML can be summarised as  "producing test set predictions for a new dataset within a fixed computational budget" \cite{NIPS2015_5872}. Essentially, the idea is to treat the process of constructing a machine learning pipeline, as a machine learning or optimisation problem itself. In doing so, the repetitive/tedious task of pipeline design is automated.

The main approaches to AutoML are TPOT \cite{Olson2016EvoBio, OlsonGECCO2016}, auto-sklearn \cite{NIPS2015_5872} and AutoWEKA \cite{thornton2013auto, kotthoff2017auto}. TPOT is based on evolutionary computation, and both auto-sklearn and AutoWEKA are based around Bayesian optimisation. TPOT and auto-sklearn both optimise scikit-learn pipelines \cite{scikit-learn}, whereas AutoWEKA optimises WEKA pipelines \cite{hall09:_weka_data_minin_softw}. All methods look to generate a machine learning pipeline by maximising some internal scorer on the training set (for example with an internal cross-validation) over a given amount of time. 

It has been shown there is little difference between the methods in terms of resulting performance \cite{gijsbers2019open, evans2019population}. We therefore choose to extend TPOT, by removing the need for specifying hyperparameter values. TPOT was selected as a base for two reasons. Firstly, TPOT features a truly flexible structure, so the discovered pipelines can be considered more "automatic" than the fixed pipeline structure of auto-sklearn (which has one feature preprocessor, one classifier, and up to three data preprocessors), which was a human-defined limit. Secondly, as TPOT uses evolutionary computation as an optimisation method, this allows for parameter-free extensions more natively.

\subsubsection{Hyperparameters}

Each AutoML comes with its own range of hyperparameters. \cref{tblParameters} summarizes the hyperparameters for TPOT and gives the default values.

\begin{table}[]
\centering
\caption{TPOT default hyperparameters}
\label{tblParameters}
\begin{tabular}{@{}ll@{}}
\toprule
                                        & \textbf{Default} \\ \midrule
Number of generations                   & 100              \\
Population size                         & 100              \\
Offspring size                          & 100              \\
Mutation rate                           & 0.9              \\
Crossover rate                          & 0.1              \\
Initial tree size                       & 1-3              \\
\textit{Optimisation metric}            & Accuracy         \\
\textit{Evaluation strategy}            & 5-fold CV        \\
\textit{Number of CPUs}                 & 1                \\
\textit{Max running time (total)}       & NA               \\
\textit{Max per model evaluation time.} & 5 Minutes        \\ \bottomrule
\end{tabular}
\end{table}

There are certain hyperparameters here which we do not automatically optimise, which are shown in \textit{italics} in the table. The reason that these hyperparameters are not optimised is because they are either regarding the availability of computational resources, or purely dependant on the desired outcome (i.e. with the scoring function). As we mention below, \cite{feurer2018towards} looks at beginning the automation of such hyperparameters via a meta-learning approach based on similar datasets.

\subsection{Parameter-Free and Adaptive Evolutionary Computation}

The proposed approach is both Adaptive (values change over time), and parameter-free (no need to specify the values).

Adaptive hyperparameters for Evolutionary Computation has seen many practical (and theoretical) works. An overview of such methods is given in \cite{aleti2016systematic} and \cite{lobo2005review}, however, most methods require the specification of additional hyperparameters, e.g., a threshold \cite{chen2009particle} or a step size \cite{thierens2002adaptive}. So rather than removing a parameter, they tend to introduce an additional parameter. Most adaptive work falls under this category as it is difficult to adapt a hyperparameter without introducing new ones (as is done here).

The importance of appropriate hyperparameter settings in EC is known \cite{lobo2007parameter, mills2015determining}, and the idea of parameter-free (or near parameter-free) optimisation has also been explored for EC \cite{randall2004near,sawai1998parameter}. However, in the context of AutoML, surprisingly there is very little work on this parameter-free optimisation. 

\citeauthor{feurer2018towards} \cite{feurer2018towards} look to automate the selection of the evaluation strategy in AutoML based on a meta-learning strategy assuming a tabular result containing results on a target metric for a set of meta-datasets. The possible evaluation strategies considered are: Holdout (67:33 train:test), k-fold Cross-validation (\textit{k}=5), and successive halving (minimum budget=16, maximum budget=256, discard ratio=2). This is only a step towards automation, as the hyper-hyperparameters of the evaluation strategy are not optimised.  That is, the train:test split of holdout is not selected, likewise, the number of folds in cross-validation is not automated, and successive halving still needs values for minimum budget, maximum budget and discard ratio to be set. This is looked at specifically in the context of auto-sklearn, although this is applicable to others (such as TPOT). In this sense, the meta-learning approach proposed in \cite{feurer2018towards} can be seen as complementary to the method proposed here (as different hyperparameters are optimised).

The goal of this work is to propose a method capable of achieving equivalent results as TPOT without requiring the need for human-defined/manually set evolutionary hyperparameters.

We achieve this by developing an adaptive and parameter-free extension of TPOT.

\section{Proposed Method}\label{secProposed}

In this section, we develop a new method for AutoML based on TPOT. The key contribution is how the search is performed. All evolutionary hyperparameters are removed and adapted at run time. These are discussed in the following section.

\subsection{Adaptive Hyperparameters}

There are several hyperparameters common to evolutionary computation methods. These define how the optimisation process should behave, and how the search space should be explored. The most common ones are the population size, crossover and mutation rates, and the number of generations. These are explored in more detail in the following sections.

\subsubsection{Population Size}

Population size ($\mu$) is an important hyperparameter for EC. A size too small may prevent the search from finding good solutions, and in the extreme case (size of 1) just behaves as hill climbing (i.e. a (1 + 1)-EA \cite{borisovsky2008comparing}). A size too large wastes computation \cite{arabas1994gavaps}, which may be better spent on further generations and can, in fact, be harmful in certain situations \cite{chen2012large}, and in the extreme case behaves as random search (if there is only time for a single generation). Somewhere between this "too small" and "too large" size is the ideal size, but of course, this is a complex choice and this depends heavily on the search space in question. Furthermore, this ideal size could change between generations \cite{eiben2004evolutionary}. If we use the evolutionary reference, we can see population growth occurs in nature and this idea of a fixed number of individuals rarely exists in natural evolution.

Here, we propose an adaptive scheme which follows the Fibonacci sequence \cite{vorobiev2012fibonacci}, i.e., $0, 1, 1, 2, 3, 5, 8, 13,  \dots$. Of course, we skip the size = 0 case, as this does not make sense for a population size. The Fibonacci sequence was chosen for a variety of reasons. It was important to follow an existing integer sequence, as we are trying to remove the need for manually specifying any hyperparameters without introducing new ones (for example a population growth rate). If we continue with the evolutionary perspective, then the Fibonacci sequence is a good candidate as it can be seen in the population growth of certain animals \cite{sigler2003fibonacci}, and appears frequently throughout nature \cite{minarova2014fibonacci, shipman2004phyllotactic}. \cite{koshy2019fibonacci} refer to the Fibonacci sequence as a "shining star in the vast array of integer sequences".  But more importantly, this sequence provides a steady increase in population size at a manageable rate, unlike say doubling each time (i.e. with exponential growth). 

Although we do not just want to grow the population size every generation, as we want the smallest population size that allows for adequate improvement. With this in mind, there are three possible cases: increase the population size, decrease the population size or maintain the population size.

\begin{itemize}
    \item  If no progression is made (maximum fitness equal to the previous maximum fitness), the population size is increased to the next number in the sequence.
    \item  If progress is made on both objectives (better score with a lower complexity), the population size is decreased to the previous number in the sequence. 
    \item If only one of the two objectives is improved (i.e. higher score with same or higher complexity, or lower complexity with the same score), the population size is maintained as we are progressing at an "acceptable" rate.
\end{itemize}

\subsubsection{Offspring Size}
In this work, we use the $(\mu + \lambda)$ EA, which is the default method in TPOT. By default, $\mu = \lambda$, with both values set to $100$, so ($100+100$).

The problem of modifying the offspring size is more complex than modifying the population size alone. However, since $\mu$ is dynamic, the choice of $\lambda$ is less important than with a fixed $\mu$. Most work on the offspring size considers the special case of $(1 + \lambda)$, where $\lambda$ is "increased roughly to the probability of producing an improving offspring" \cite{jansen2005choice}. \citeauthor{jansen2005choice} \cite{jansen2005choice} also claim that a dynamic size "might lead to a performance similar to that of a $(1 + \lambda)$ EA with an optimal value of $\lambda$ without having to determine that optimal value a priori", which is the goal we aim to achieve for all hyperparameters in this work.

In the previous subsection, we increase $\mu$ based on past performance. We then fix the value of $\lambda$ to be the difference between the current number $fib(i)$ and the next number $fib(i+1)$ in the Fibonacci sequence, $\lambda=fib(i+1)-fib(i)$. For example, if we have a population size of $\mu=144$, the offspring size would  $\lambda=89$. Since $\mu$ is dynamic (unlike the theoretical cases of $\mu=1$), there is little extra knowledge we can use for specifying $ \lambda$, which is why we reuse the knowledge from selecting $\mu$.

If $\mu_{gen+1} > \mu_{gen}$, the result will be that all the offspring are used (i.e. no selection performed). If $\mu_{gen+1} <= \mu_{gen}$, then only the best $\mu_{gen+1}$ will be selected from $\mu + \lambda$. In all cases, $\lambda$ is positive so a generation will always produce offspring.

\subsubsection{Crossover, Mutation, and Reproduction}

In many EC methods, the sum of the crossover, mutation and elitism rates should be 1. However, in this case, since we are using the $\mu + \lambda$ algorithm, the elitism is implicit.  The best $\mu$ individuals will be kept from $\mu + \lambda$, so top performers are never lost.

We are left with the crossover and mutation rates, which should sum to 1.

Of course, the rates need to begin somewhere. Since our first population size is 1, the mutation rate begins at 1 and the crossover at 0. The reason for this is we can not perform crossover with a single individual, so this is the only option.

For subsequent generations, the rates become adaptive. We do not introduce any new hyperparameters (as this would defeat the parameter-free idea) such as a decay rate, or the number of generations without improvement, instead, we base the rates purely on statistics about the population.

Mutation randomly changes an individual in an effort to promote diversity. Therefore, in populations with low diversity, mutation should be performed frequently (as crossing over two structurally similar). As a "naive" measure of diversity, we use the standard deviation $\sigma$ of the population's fitness. Since this is multiobjective (maximise performance and minimise complexity), we only use the `performance' measure here and not both objectives, in this case, $F_{1}$-score when discussing fitness. This is a fast and approximate measure of diversity, but we call this naive as it is not necessarily a measure of behavioural or structural diversity. In future work, we would like to explore the idea of semantically, rather than just "fitness" diversity.

With $\sigma$ as a "diversity" measure (although strictly this is just a fitness diversity measure), we use this to set the mutation rate. A high $\sigma$ indicates high (fitness) diversity, so a lower mutation rate can be used. Likewise, a low $\sigma$ indicates similar performance, so a higher mutation rate should be used. The equation for dynamically setting the mutation rates is given in \cref{eqMutation}, where $\sigma_{gen}$ represents the standard deviation of the population's fitness in generation $gen$.

The maximum value for $\sigma$ is $0.5$, since the objective is bound between $[0,1]$. To standardise this, and ensure the mutation rate can be set to a value above $0.5$, we scale $\sigma$ by the maximum observed standard deviation from all previous generations. We scale by the maximum observed standard deviation rather than the maximum theoretical standard deviation (of 0.5) for two reasons. Firstly, the mutation should be based on the current problem, in certain tasks there may naturally be a very small variation in results (i.e. simplistic problems where all methods perform well, or problems with local optima where many methods can hit but not surpass), and crossover would never be trialled. Instead, if the mutation is relative to previous observations (rather than the theoretical maximum variation), crossover can occur if comparatively this generation seems diverse.  Secondly, this is more general in the sense that it will work for unbounded objective measures with no theoretical maximum.

\begin{equation}\label{eqMutation}
    mutation_{gen} = 1 - \frac{\sigma_{gen}}{ max\{ \sigma_g  : g = 1 \dots gen \} }
\end{equation}

As the rates should sum to 1, the crossover rate becomes the remainder ($1 - mutation$). In this sense it is not possible to vary one without adjusting the other, so attempting to adjust both would be redundant.

\subsubsection{Maximum Generations} 
With an adaptive population size, the need for a certain number of generations becomes less important/meaningful. Therefore, this can be easily removed, and instead, we just replaced with a running time (in minutes). An alternative would be a certain number of evaluations (i.e. 10,000 models), but a maximum running time seems more user-friendly due to the large variation in the cost of evaluation for different models. For this reason, we remove the number of generations and replace with a maximum number of minutes to allow for optimisation. Of course, this running time can be considered a parameter, but the running time tends to be an easy choice based on computational resources.

\subsubsection{Additional Hyperparameters}

There are also some additional hyperparameters which we remove. These are "internal" hyperparameters, not user-specified in TPOT, but nevertheless the removal is another step towards the goal of full automation.

The first is the size of the initial pipelines, which is set by default to be between 1 and 3 nodes. This is instead replaced by stumps, so pipelines begin as a single node (a classifier).

Second is related to the genetic operators (crossover and mutation). By default, these are tried up to 50 times to generate a unique offspring. This is replaced by a "diversity" preserving crossover and mutation, which considers only unexplored offspring directly (i.e. those not in the existing cache). As a result, the possible mutation operators can be seen as dynamic and will be disabled for particular individuals if they cannot result in a novel individual.

\subsubsection{Manual Hyperparameters}
Currently, there are hyperparameters which still exist around computational power and scoring. For example, what metric should we use to measure performance? Accuracy, $F_{1}$-score, or a weighted accuracy with class importances? We did not look to remove this, as this is entirely dependent on the problem, and users desire. For example, we may have a binary problem where class 1 is far more important to predict than class 2 (say cancer diagnosis), but this can not be inferred from the data. Likewise, other settings such as max computational time, and number of CPUs to use remains as a user-specified parameter, as this depends on the computational resources available.  Future work could look at predicting the performance based on the resources given, based on meta-learning on similar datasets (i.e. treat the selection as a regression problem), but this would be a very broad approximation and not something we attempted in this work.

\subsection{Algorithm}

While the representation of pipelines (trees) and original search space maintains the same as TPOT (overview given in \cref{secBackground} and full description in  \cite{OlsonGECCO2016}), the key development is how this search space is explored.

Rather than hyperparameters being fixed and specified a priori, they are adapted at run time by the algorithm given in Algorithm \ref{algPseudoCode}.

\begin{algorithm}[!ht]
\SetAlgoLined

\SetKwProg{Def}{def}{:}{}

\Def{adaptiveEa(run\_time: int)}{
    population\_size = 1\;
    population = [random\_stump()]\;
    evaluate(population)\;

    \While{time $<$ run\_time}{
        offspring\_size = preceding\_fibonacci(population\_size)\;

        offspring = apply\_genetic\_operations(population, offspring\_size, mutation\_rate)\;

        evaluate(offspring)\;

        \If{improved both objectives}{
            population\_size = preceding\_fibonacci(population\_size)\;
        }
        \ElseIf{improved one objective}{
            population\_size = population\_size\;
        }
        \Else{
            population\_size = proceeding\_fibonacci(population\_size)\;
            mutation\_rate = 1 - (fitness\_std / max\_std )\;
        }

        NSGA\_II(population + offspring, population\_size)\;
    }

    return population\;
}
\caption{Pseudo Code for adaptive evolution}
\label{algPseudoCode}
\end{algorithm}

\section{Results}\label{secResults}

\subsection{Comparisons}
While it would be ideal to perform a grid search to find ideal evolutionary hyperparameters for each dataset for comparison, this is computationally infeasible. Instead, we compare to the default values (as this is what will be most commonly used), which we assume are already the result of large scale exploration of such values.

The default values for TPOT are given in \cref{tblParameters}. 
The default values from TPOT are initially surprising to someone familiar with GP, which helps reaffirm that the setting of such values can be difficult. As not only are you required to know how EC methods work and how these values affect the search space, but you must also know about the search space of machine learning pipelines (huge and plagued by local optima). 

For the proposed method, this does not require specifying the hyperparameters. The method was outlined in detail in \cref{secProposed}.

Both methods were capped at a 1-hour run-time due to computational restrictions.

\subsection{Datasets}

The datasets proposed in \cite{gijsbers2019open} as "An Open Source AutoML Benchmark" were used for comparison here. In total there are 39 datasets proposed for the benchmark, however, 9 of these did not generate results within the allowed computational budget, so were excluded from the results. The datasets were chosen by \citeauthor{gijsbers2019open} to be of sufficient difficulty, be representative of real-world problems, and have a large diversity of domains.

\subsection{Statistical Tests}

To compare the methods we generate an average (mean) performing using 10-fold cross-validation (given in \cref{tblResults}). General significance testing is performed using a Wilcoxon signed-rank test. We call this general significance testing as comparisons are made across datasets (rather than across individual runs on each dataset). Specifically, this type of significance testing avoids the often inflated Type-1 errors associated with repeatedly performing significance tests across many datasets (as is more commonly seen, but we believe is problematic and according to \cite{demvsar2006statistical} should not be used). $\alpha=0.05$ is used for reporting significance, with the \textit{p}-values also given in \cref{tblResults}. This follows the suggestions in both \cite{demvsar2006statistical, garcia2008extension} for fair classifier comparison, in an effort to remedy the "multiple comparisons problem" which often arises in statistical comparisons \cite{sullivan2016common} (particularly in cases such as this with a large number of datasets, meaning a large number of comparisons being performed). Another benefit this approach has over the more typically seen multiple comparisons, is it allows us to draw more general conclusions about the performance of the methods.

We report the average $F_{1}$-score, average resulting complexity, and also since the method is multiobjective, we compute the average hypervolume \cite{fonseca2006improved} as well for comparing the frontiers. The reference point used in all cases was (0,10), meaning a $F_{1}$-score of zero and a complexity of 10 (this is the approximate nadir point, i.e. the worst  value of each objective in the Pareto optimal set \cite{deb2006towards}). A complexity of 10 was chosen as no individuals in any frontiers had a complexity $\geq$ 10, so even though complexity is not bounded to be $<$ 10, this was a fair choice. $F_{1}$-score was chosen over accuracy as we cannot assume an equal distribution of class instances across the datasets.

\subsection{Discussion}

The average results and significance testing is presented in \cref{tblResults}. We can see there is no statistically significant difference between the proposed automated method and the human-expert configured baseline, therefore, in general, we can conclude that the proposed method is able to achieve equivalent performance to the human selected baseline.

When considering individual runs, we can see the results tend to be very similar. In rare cases, there are large differences, but these go both ways and appear to cancel out (i.e. one method is never always better than the other), confirmed with the general significance testing. An example can be seen with the segment dataset where the baseline gets ~94 and the proposed gets ~84, but then on the covertype dataset the baseline gets ~67 and the proposed gets ~87. 

\begin{table}[]
\caption{Average results. The proposed method comes under the \textbf{Automatic} column, and TPOT with the default (human specified) hyperparameter values comes under the Human column. $F_{1}$-score should be maximised (higher the better), with an optimal value of 100 (scaled from $0\dots1$ to $0\dots100$ for readability). The complexity should be minimised (lower the better), with an optimal value of 1. The hypervolume should also be maximised (higher the better). \textit{p}-values are presented in the final row. }
\label{tblResults}
\resizebox{\columnwidth}{!}{%
\begin{tabular}{@{}lcccccc@{}}
\toprule
                                            & \multicolumn{2}{c}{$F_{1}$-Score}                       & \multicolumn{2}{c}{Complexity}                     & \multicolumn{2}{c}{Hypervolume}                    \\ \midrule
                                            & Human             & \textbf{Automatic}             & Human             & \textbf{Automatic}             & Human              & \textbf{Automatic}            \\ \midrule
adult                                       & 87.10             & 86.99                          & 2.00              & 2.00                           & 7.84               & 7.83                          \\
airlines                                    & 63.54             & 57.80                          & 1.00              & 2.40                           & 5.72               & 5.20                          \\
anneal                                      & 99.62             & 93.86                          & 1.70              & 1.50                           & 8.97               & 8.46                          \\
australian                                  & 87.52             & 87.81                          & 4.70              & 2.90                           & 7.97               & 7.89                          \\
bank                              & 85.94             & 86.06                          & 2.00              & 2.20                           & 7.76               & 7.77                          \\
blood            & 75.92             & 75.07                          & 3.30              & 2.20                           & 6.93               & 6.76                          \\
car                                         & 96.52             & 91.93                          & 4.50              & 1.90                           & 8.86               & 8.30                          \\
christine                                   & 73.93             & 69.99                          & 1.00              & 1.30                           & 6.65               & 6.29                          \\
cnae-9                                      & 96.30             & 95.75                          & 1.80              & 2.30                           & 8.69               & 8.63                          \\
covertype                                   & 67.51             & 87.24                          & 1.43              & 1.00                           & 6.68               & 7.85                          \\
credit-g                                    & 72.79             & 75.11                          & 3.10              & 2.20                           & 6.72               & 6.92                          \\
dilbert                                     & 96.84             & 59.62                          & 1.10              & 3.00                           & 8.72               & 5.34                          \\
dionis                                      & 21.02             & 20.84                          & 1.00              & 1.00                           & 1.89               & 1.88                          \\
fabert                                      & 69.77             & 68.41                          & 1.89              & 1.43                           & 6.29               & 6.18                          \\
fashion                               & 56.45             & 58.36                          & 1.00              & 3.00                           & 5.08               & 5.23                          \\
guillermo                                   & 66.47             & 75.80                          & 1.00              & 1.00                           & 5.98               & 6.82                          \\
helena                                      & 25.48             & 21.26                          & 1.43              & 1.00                           & 2.27               & 1.91                          \\
higgs                                       & 72.22             & 71.36                          & 1.00              & 1.22                           & 6.50               & 6.42                          \\
house\_16H                                  & 5.56              & 4.43                           & 1.83              & 1.00                           & 0.50               & 0.40                          \\
jannis                                      & 70.27             & 69.84                          & 1.40              & 1.90                           & 6.31               & 6.26                          \\
jasmine                                     & 81.07             & 81.72                          & 1.75              & 3.70                           & 7.33               & 7.41                          \\
jungle & 82.65             & 84.11                          & 1.70              & 1.50                           & 7.63               & 7.59                          \\
kc1                                         & 82.48             & 82.74                          & 2.90              & 1.50                           & 7.47               & 7.51                          \\
kr-vs-kp                                    & 99.34             & 99.47                          & 2.80              & 3.00                           & 8.96               & 8.96                          \\
mfeat                               & 97.70             & 97.15                          & 1.60              & 1.60                           & 8.80               & 8.74                          \\
miniboone                                   & 93.33             & 93.76                          & 1.10              & 1.90                           & 8.40               & 8.44                          \\
nomao                                       & 97.03             & 96.81                          & 1.12              & 2.20                           & 8.73               & 8.72                          \\
numerai                                     & 51.90             & 51.66                          & 1.50              & 1.00                           & 4.67               & 4.65                          \\
phoneme                                     & 91.29             & 90.00                          & 2.90              & 2.30                           & 8.23               & 8.17                          \\
riccardo                                    & 77.63             & 98.79                          & 1.56              & 1.00                           & 6.96               & 8.89                          \\
robert                                      & 41.34             & 36.43                          & 2.00              & 1.50                           & 3.67               & 3.28                          \\
segment                                     & 93.97             & 84.41                          & 2.60              & 2.10                           & 8.52               & 7.61                          \\
sylvine                                     & 95.88             & 95.84                          & 2.90              & 3.60                           & 8.62               & 8.64                          \\
volkert                                     & 63.72             & 54.56                          & 1.00              & 1.10                           & 5.73               & 4.91                          \\ \midrule
\textbf{Significant}                        & \multicolumn{2}{c}{\cellcolor[HTML]{FFCCC9}\textit{p}=0.1218} & \multicolumn{2}{c}{\cellcolor[HTML]{FFCCC9}\textit{p}=0.7187} & \multicolumn{2}{c}{\cellcolor[HTML]{FFCCC9}\textit{p}=0.0502} \\ \bottomrule
\end{tabular}
}
\end{table}

\subsubsection{Frontiers}

\begin{figure*}
    \centering
    \includegraphics[width=\textwidth
    ]{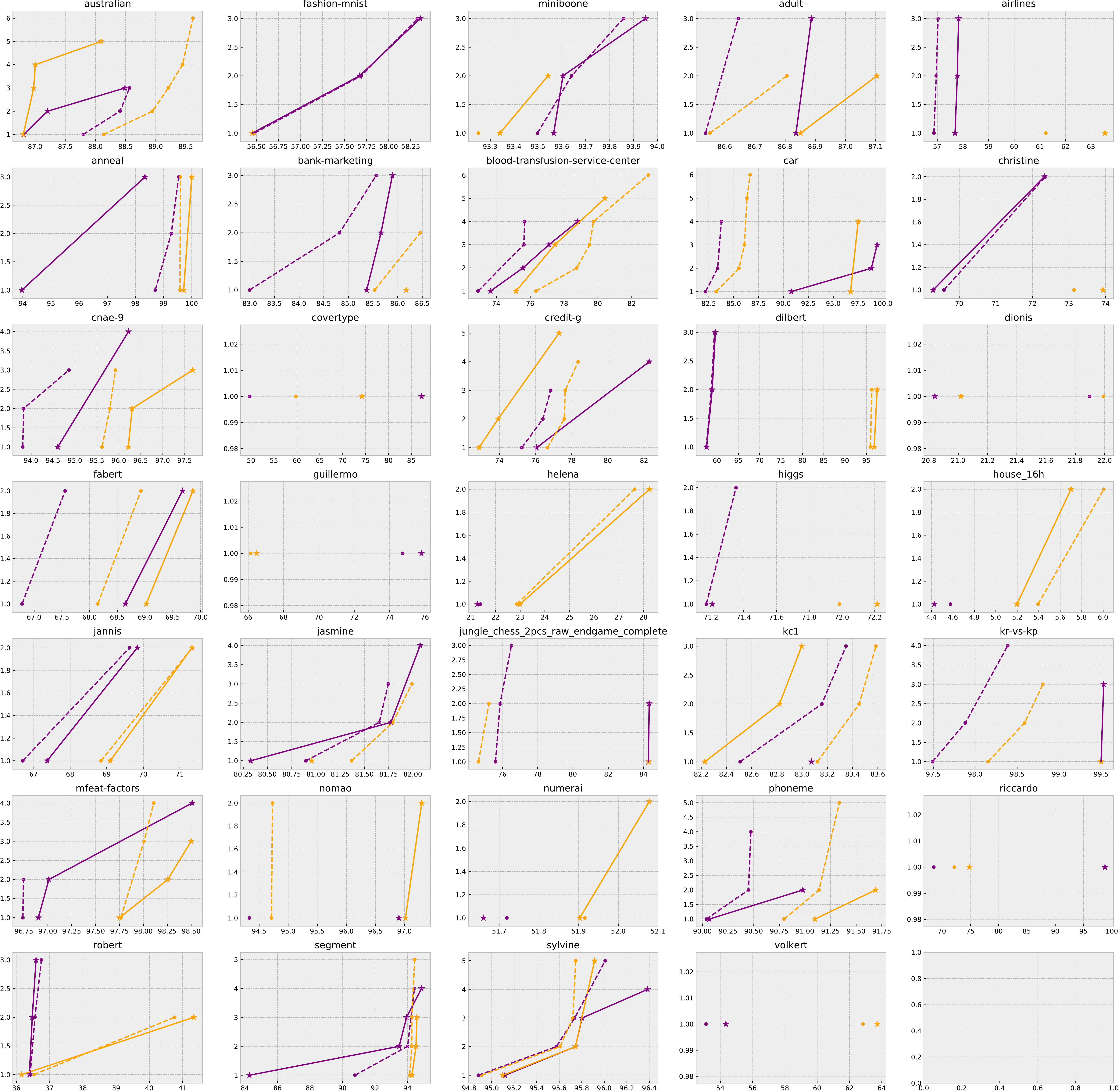}
    \caption{Average frontiers. The proposed method is in \textcolor{violet}{purple}, and the comparison method TPOT is in \textcolor{orange}{orange}. The training frontier is given as dashed lines, and the true testing frontier given as a solid line. The training points in the frontier are indicated with a `.', and the testing points indicated with a `*'.  The ideal position is the bottom right (i.e. a score of 100 on the x-axis, and a complexity of 1 on the y-axis)}
    \label{figFrontiers}
\end{figure*}

We visualise the frontiers for comparison in \cref{figFrontiers}. Since the results are averaged over several runs to produce a reliable result, the frontiers visualised are themselves averaged for a result. 

As each run may result in a different size frontier (for example a different number of complexities), we can not just do a pairwise average of the frontier for each method on each dataset.  Instead, for visualisation, we compute the average resulting score for each complexity and then remove points that were dominated from this set to produce a non-dominated front. This was only done for visualisation sake, and not for the significance testing above.

When viewing the frontiers, there is no clear winner between the methods (again confirmed with the statistical testing done above on the resulting hypervolumes). No drastic overfitting occurs with either of the methods (no large difference between training and test performance). 

\section{Further Analysis}

To get a further understanding of how the method works, we analyse each of the adaptive hyperparameters.

\begin{figure*}
    \centering
    \begin{subfigure}[b]{\textwidth}
    \includegraphics[width=\textwidth]{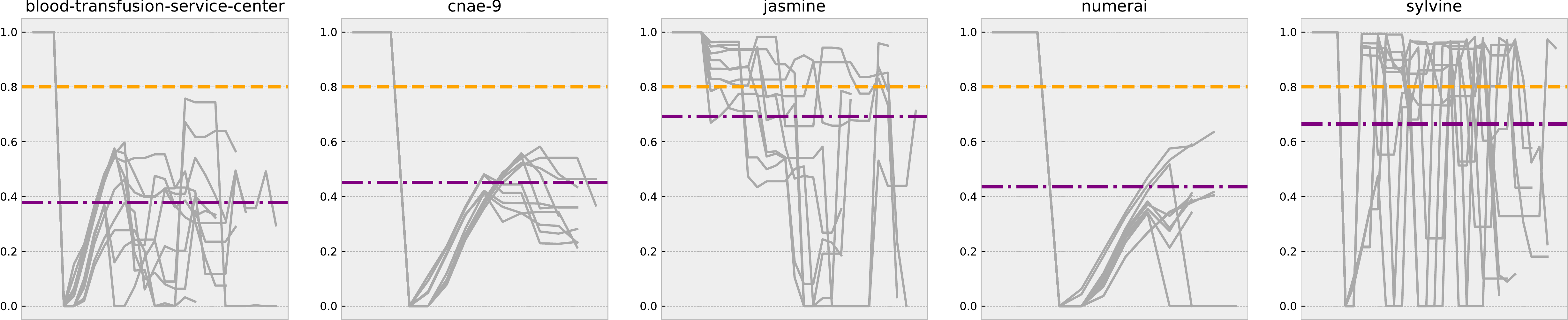}
    \caption{Mutation rate.}
    \label{figMutation}
    \end{subfigure}

    \begin{subfigure}[b]{\textwidth}
    \includegraphics[width=\textwidth]{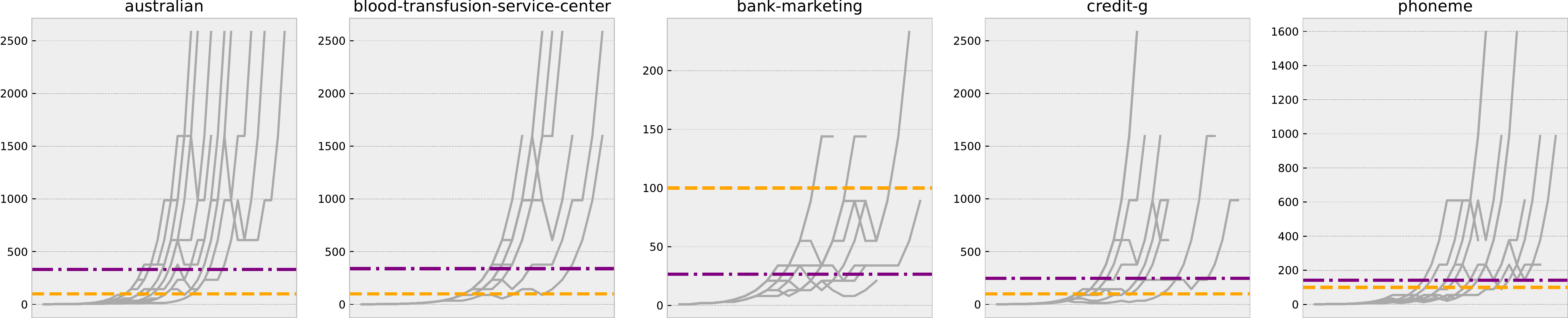}
    \caption{Population size.}
    \label{figPopulation}
    \end{subfigure}
    
    \caption{Adaptive hyperparameter (y) over generations (x). Each grey line indicates the result from a single run. The dotted \textcolor{orange}{orange} line indicates the default (expert chosen) value in TPOT. The dotted \textcolor{violet}{purple} line shows the average rate for the adaptive method.}
\end{figure*}

\subsection{Adaptive Hyperparameters}

\subsubsection{Mutation}

The adaptive mutation rates are shown in \cref{figMutation}. Only 5 datasets are chosen for comparison due to space restrictions, however, these are generally representative of the behaviours.

The mutation rates start at 1 since the population size begins at 1 and crossover cannot be applied.  We see a large variation in the mutation rates, both across generations for a dataset and between datasets. This indicates the adaptations are serving their purpose, as the values are different throughout time and between datasets, indicating no one universal optimal value. 

There are no clear trends in how the mutation behaves between datasets, showing the rates are very much problem dependant. One commonly believed/argued point is that mutation should decrease over time, meaning we start with a high mutation (and high exploration) and over time we begin to focus more on exploitation, although it is worth mentioning with genetic programming it is difficult to decide whether an operation is promoting exploration or exploitation since both crossover and mutation can be argued either way.  This appears to be roughly what happens on the jasmine dataset (with high fluctuations but mutation rate trending downward), whereas on the cnae-9 and numerai datasets we see the opposite trend. On blood, cnae-9, and numerai the rate starts high, drops quite drastically (indicating a diverse population was generated), then begins an upward trend. On the sylvine dataset, there is no clear trend at all. Changes appear to occur between the extremes, going from near zero to near one, indicating each time after a population with poor diversity, a highly diverse population was then generated.

Reassuringly, across individual runs for a given dataset, we see similar patterns in how the mutation rate changes. For example, the cnae-9 dataset shows this clearly. This is a desirable property as with slight variations in the training data (i.e. a new fold in 10-fold CV), the evolutionary process shouldn't be drastically different since we are trying to optimise for some true population and not just the training data. This is confirmed in the results (\cref{secResults}), which show we do not overfit to the training data, which is very important with such adaptive methods.

The average mutation rate (indicated by the dashed purple line in \cref{figMutation}) is generally lower than the default human chosen TPOT rate, which is a relatively high rate of 0.9. We still find the average mutation rate to be much higher than typical values for GP, perhaps due to the very large search space which is also plagued with local optima, meaning high levels of exploration is desirable.

\subsubsection{Population Size}

The growth of the population size is visualised in \cref{figPopulation}. Again, only 5 datasets are used for visualisation purposes. 

Unlike the mutation rates, the population path is much more well behaved. This is not by chance, but instead because the Fibonacci sequence is followed -- which explains the curves seen. We can see that the population size is very rarely stagnant, and does not just increase monotonically. Often the size will be maintained or drop. However, it does appear relatively uncommon to decrease for sustained periods  (indicating difficulty in optimisation, requiring a larger population size). 

Interestingly, we can observe the average population size (purple dotted line in \cref{figPopulation}) is often very close to the default setting of the expert method of 100 individuals (orange dotted line). However, this may just be a factor of the allowable running time, as the common trend between datasets seems to be for the population to grow as the evolution progresses.

Again, for an individual dataset, the individual runs (dark grey lines), all appear to behave similarly to each other. This shows robustness in the adaptations, which is a desirable trait. Between datasets, the population size trends upwards (as it becomes harder and harder to make progress), but the graphs may be somewhat deceiving at first glance since between datasets we can see the y-axis can operate on different scales. For example, bank-marketing gets to a peak of around 200, whereas blood-transfusion is around 2500. This is a factor of the size of the datasets, where blood-transfusion has ~45000 instances and 17 features, whereas the bank-marketing has ~750 instances and 5 features. Again, this is a beneficial trait as it shows robustness to various dimensionalities and number of training instances.

\section{Conclusions and Future Work}

In this work, we proposed a parameter-free approach to AutoML. The search process begins with a single randomly chosen estimator and automatically evolves to a well-performing pipeline without the need for specifying any evolutionary hyperparameters such as population size, crossover rate, or mutation rate. We evaluated the proposed method across a large number of datasets and found the results to equivalent to the current state-of-the-art approach which uses human knowledge for selecting hyperparameters. This is encouraging as it demonstrates a step towards true "Automation", in AutoML. This opens further research into whether entirely self-trained systems can see improvements over human-assisted approaches like has been seen in other areas such as gameplay \cite{silver2017mastering}. Future work in the area of AutoML can focus on trying to push the field towards complete automation.

Here we looked at adapting the search hyperparameters throughout the optimisation process. An additional area of future research could also look at adapting the search space itself, where unpromising nodes or subtrees are removed entirely. This is difficult, as subtrees may only be poor performing in their current context (i.e. ensembling "poor" classifiers can improve performance if they are diverse \cite{schapire1990strength}). 

This work sets the foundation for fully parameter-free automated approaches to AutoML.

\bibliographystyle{IEEEtranN}
\bibliography{bib}

\end{document}